\documentclass{article} 
\usepackage{iclr2025_conference,times}

\usepackage{amsmath,amsfonts,bm}

\def\eqref#1{equation~\ref{#1}}

\def\1{\bm{1}}

\def\rve{{\mathbf{e}}}

\def\rvx{{\mathbf{x}}}

\def\vtheta{{\bm{\theta}}}

\def\vm{{\bm{m}}}

\def\vp{{\bm{p}}}

\def\vu{{\bm{u}}}

\def\vx{{\bm{x}}}

\def\mQ{{\bm{Q}}}

\DeclareMathAlphabet{\mathsfit}{\encodingdefault}{\sfdefault}{m}{sl}
\SetMathAlphabet{\mathsfit}{bold}{\encodingdefault}{\sfdefault}{bx}{n}

\newcommand{\E}{\mathbb{E}}

\newcommand{\KL}{D_{\mathrm{KL}}}

\usepackage{url}
\usepackage{xspace}
\usepackage{mathtools}
\usepackage{booktabs}
\usepackage{multirow}
\usepackage{wrapfig}
\usepackage{xcolor}
\usepackage{caption}
\usepackage{algorithm}
\usepackage{algpseudocode}
\usepackage{algorithmicx}

\definecolor{bleudefrance}{rgb}{0.19, 0.55, 0.91}
\usepackage[colorlinks=true,linkcolor=orange,citecolor=bleudefrance]{hyperref}

\title{Beyond Autoregression: Discrete Diffusion for Complex Reasoning and Planning}

\author{%
Jiacheng Ye$^1$, Jiahui Gao$^2$
  Shansan Gong$^1$,
  Lin Zheng$^1$ \\ \bf Xin Jiang$^2$, Zhenguo Li$^2$, Lingpeng Kong$^{1}$ \\
  $^1$ The University of Hong Kong \quad $^2$ Huawei Noah's Ark Lab \\ 
  \texttt{\small carsonye@connect.hku.hk}
}
\newcommand{\ourmodel}{\textsc{MGDM}\xspace}
\newtheorem{proposition}{Proposition}
\newcommand{\markred}{\textcolor[HTML]{E06666}}

\newcommand{\reb}[1]{#1}

\iclrfinalcopy 
\begin{document}
\maketitle

\begin{abstract}
Autoregressive language models, despite their impressive capabilities, struggle with complex reasoning and long-term planning tasks. We introduce discrete diffusion models as a novel solution to these challenges. Through the lens of subgoal imbalance, we demonstrate how diffusion models effectively learn difficult subgoals that elude autoregressive approaches. We propose Multi-Granularity Diffusion Modeling (\ourmodel), which prioritizes subgoals based on difficulty during learning. On complex tasks like Countdown, Sudoku, and Boolean Satisfiability Problems, \ourmodel significantly outperforms autoregressive models without using search techniques. For instance, \ourmodel achieves 91.5\% and 100\% accuracy on Countdown and Sudoku, respectively, compared to 45.8\% and 20.7\% for autoregressive models. Our work highlights the potential of diffusion-based approaches in advancing AI capabilities for sophisticated language understanding and problem-solving tasks. All associated codes are available at \href{https://github.com/HKUNLP/diffusion-vs-ar}{https://github.com/HKUNLP/diffusion-vs-ar}.
\end{abstract}
\section{Introduction}

In recent years, autoregressive language models (LMs; \citealt{bengio2000neural}) have dominated the landscape of natural language processing and artificial intelligence. Empowered by scaling laws \citep{kaplan2020scaling}, these models have demonstrated impressive performance across various applications~\citep[\emph{inter alia}]{chatgpt,achiam2023gpt,claude,team2023gemini}. However, this apparent success masks significant limitations that are becoming increasingly evident. Autoregressive models inherently struggle with tasks requiring complex reasoning, long-term planning, and maintaining global coherence~\citep{bubeck2023sparks, valmeekam2023planning,valmeekam2024planbench,dziri2024faith,kambhampati2024llms}. These shortcomings represent substantial challenges in developing AI systems capable of robust problem-solving and adaptable cognition~\citep[\emph{inter alia}]{wu2022autoformalization,zhao2023decomposing,trinh2024solving,yao2023react,shinn2024reflexion}. While autoregressive approaches have driven considerable progress, their limitations suggest that they may not be the optimal solution for all aspects of machine intelligence. As the field evolves, it becomes increasingly important to explore alternative paradigms that can address these inherent drawbacks and potentially offer new avenues for advancement in AI capabilities.

In response to these limitations, recent research has focused on addressing the inherent constraints of autoregressive models. Various strategies have been explored, including the integration of search algorithms at inference~\citep{yao2024tree,besta2024graph} and the incorporation of backtracking supervision during training~\citep{lehnert2024beyond,gandhi2024stream}. However,  these approaches are not without their own drawbacks: the former often incurs significant computational costs, while the latter frequently results in verbose inputs and suboptimal performance.

To address this challenge, we argue for a fundamentally different modeling approach: discrete diffusion models. While most contemporary language models are autoregressive, diffusion-based models have become predominant in image~\citep{dhariwal2021diffusion,rombach2022high,peebles2023scalable} and video domains~\citep{ho2022video,wu2023tune,videoworldsimulators2024}. Diffusion models are also gaining traction in various other applications, such as protein desiging~\citep{xu2022geodiff,hoogeboom2022equivariant,corso2023diffdock} and planning in reinforcement learning~\citep{janner2022planning,ajay2022conditional,chi2023diffusion}. In this work, we reveal that discrete diffusion models demonstrate significantly superior performance compared to the autoregressive counterparts, particularly in tasks requiring complex planning and reasoning.

To substantiate this argument, we first examine the problem through the lens of \emph{subgoal imbalance} (\S\ref{sec:planning_task}). We present both theoretical and empirical evidence via a synthetic planning task (Figure~\ref{fig:findpath}) to illustrate why autoregressive models struggle with these types of problems, often achieving near-random performance. In contrast, we demonstrate how diffusion models effectively learn the subgoals that challenge autoregressive models (\S\ref{sec:how-diffusion}). The key insight lies in the training objective of diffusion models, where difficult subgoals are decomposed into a diverse range of interrelated views within a multi-view learning framework~\citep{xu2013survey}. Each of these views is more manageable, resulting in an overall easier and more effective learning process.

Building upon these insights, we propose a natural extension to current discrete diffusion models, which we term multi-granularity diffusion modeling (\ourmodel; \S\ref{sec:mdm}). This approach prioritizes different subgoals based on their difficulty during the learning process, leading to more effective learning outcomes and faster convergence.

In our experimental evaluation (\S\ref{sec:experiments}), we focus on substantially more complex problem-solving tasks, such as Countdown \citep{gandhi2024stream} and Sudoku \citep{sudoku}. These tasks demand extensive planning over a large number of combinations and pose challenges even for commercial Large Language Models (e.g., GPT-4 \citealt{achiam2023gpt}). Notably, without employing any search techniques, \ourmodel achieves 91.5\% and 100\% accuracy on Countdown and Sudoku respectively, while its autoregressive counterpart only solves 45.8\% and 20.7\% of the problems. Additionally, we conduct experiments on the Boolean Satisfiability Problem (SAT), an NP-complete problem \citep{cook1971complexity} that represents a wide range of constraint satisfaction problems. Our model exhibits superior performance in solving SAT problems with higher accuracy compared to the autoregressive alternative, particularly when dealing with an increased number of variables and constraints. Through this systematic exploration, we aim to demonstrate the potential advantages of diffusion-based approaches in addressing sophisticated language understanding and generation challenges.

\section{Background}
\label{sec:bg}
\subsection{Auto-regressive Modeling}
Let $\rvx \coloneq (\vx_1,\dots,\vx_N)$ denote a sequence drawn from a data distribution $q(\rvx)$. For decades, it has been common to factorize the joint probabilities of a sequence of tokens as the product of conditional probabilities~\citep{jelinek1980interpolated,bengio2000neural}:
\begin{equation}
p_\vtheta(\rvx)=p_\vtheta(\vx_1)\prod_{n=2}^{N}p_\vtheta(\vx_n \mid \vx_{1:n-1}),
\end{equation}
where $\vtheta$ parameterizes the model distribution and $\vx_{1:n-1}\coloneqq \vx_1, \dots, \vx_{n-1}$. In order to optimize the generative model $p_\vtheta(\rvx)$ to fit the data distribution $q(\rvx)$, we
optimize the negative log-likelihood:
\begin{equation}
L_{\text{AR}} = -\E_{q(\rvx)}\log p_\vtheta(\rvx)=-\E_{q(\rvx)}\sum_{n=1}^N \log p_\vtheta(\vx_n \mid \vx_{1:n-1}).
\label{eq:arloss}
\end{equation}

\subsection{Discrete Diffusion Modeling}
 Discrete diffusion models~\citep{sohl2015deep,hoogeboom2021argmax,austin2021structured} are a class of latent variable models characterized by a forward noising process and a learned reverse denoising process. The forward process $q(\rvx_{1:T}|\rvx_0) = \prod_{t=1}^T q(\rvx_t|\rvx_{t-1})$ corrupts the original data $\rvx_0 \coloneqq \rvx$ into a sequence of increasingly noisy latent variables $\rvx_{1:T}\coloneqq \rvx_1, \dots, \rvx_T$. The backward process learns to gradually denoise the latent variables
to the data distribution given by:
\begin{equation}
\label{diffu_rev}
    p_{\vtheta}(\rvx)=\sum_{\rvx_{1:T} \sim q}p(\rvx_T)\prod_{t=1}^Tp_{\vtheta}(\rvx_{t-1}|\rvx_t).
\end{equation}
Due to the intractable marginalization, we typically
optimize a variational upper bound on the negative log-likelihood:
\begin{align}
    L_{\mathrm{DM}} = \E_{q(\rvx_0)}\bigg[&
       \underbrace{D_{\mathrm{KL}}[q(\rvx_T | \rvx_0) \vert\vert p(\rvx_T)]}_{L_T}
    + \sum_{t=2}^T \underbrace{\mathbb E_{q(\rvx_t|\rvx_0)} \big[
        D_{\mathrm{KL}}[q(\rvx_{t-1} | \rvx_t, \rvx_0) \vert\vert 
        p_{\vtheta}(\rvx_{t-1}|\rvx_t)]
        \big]}_{L_{t-1}} \nonumber \\[-0.5em]
      &\underbrace{- \mathbb E_{q(\rvx_1|\rvx_0)} [\log p_{\vtheta}(\rvx_0|\rvx_1)]}_{L_0}
    \bigg],
\label{eq:dm_original}
\end{align}
where $L_T$ is a constant when one uses a fixed prior $p(\rvx_T)$.
By defining both the forward and backward distribution as categorical distribution, e.g., $q(\rvx_t|\rvx_{t-1})= \mathrm{Cat}(\rvx_t;\vp = \mQ_t^\top \rvx_{t-1})$ where $\mQ_t$ is a pre-defined  $K\times K$ transition matrix and $K$ is the size of categories, and $p_{\vtheta}(\rvx_{t-1}|\rvx_t)=q(\rvx_{t-1} | \rvx_t, f(\rvx_t;\vtheta))$, the forward process posterior $q(\rvx_{t-1} | \rvx_t, \rvx_0)$ and each KL term can be calculated analytically~\citep{hoogeboom2021argmax,austin2021structured}.

\section{Subgoal Imbalance and Multi-granularity Diffusion Models}
\label{sec:subgoal}
In this section, we employ a motivation example (\S\ref{sec:planning_task}) to elucidate the challenges faced by autoregressive models in specific scenarios. Through this analysis, we introduce the concept of \emph{subgoal imbalance}—wherein some subgoals are inherently more difficult than others—which offers insights into these difficulties. We then extend our discussion in \S\ref{sec:how-diffusion} to examine how diffusion models can more effectively address and learn these \emph{hard subgoals}, effectively overcoming the limitations observed in autoregressive approaches. We finally propose Multi-Granularity Diffusion Modeling (\ourmodel; \S\ref{sec:mdm}) as a natural extension of discrete diffusion models to better address these challenges and improve performance on complex tasks requiring planning and reasoning.

\subsection{Subgoal Imbalance in Autoregressive and Diffusion Modeling}
\label{sec:planning_task}

\begin{figure}[t]
  \centering
  
\begin{minipage}{.33\linewidth}
\centering

\includegraphics[width=0.95 \textwidth]{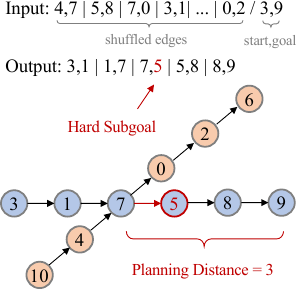}
\vspace{-5.5pt}
\caption{The planning task.}
\label{fig:findpath}

\end{minipage}
\begin{minipage}{.65\linewidth}
\vspace{10pt}
\centering
\includegraphics[width=0.99 \textwidth]{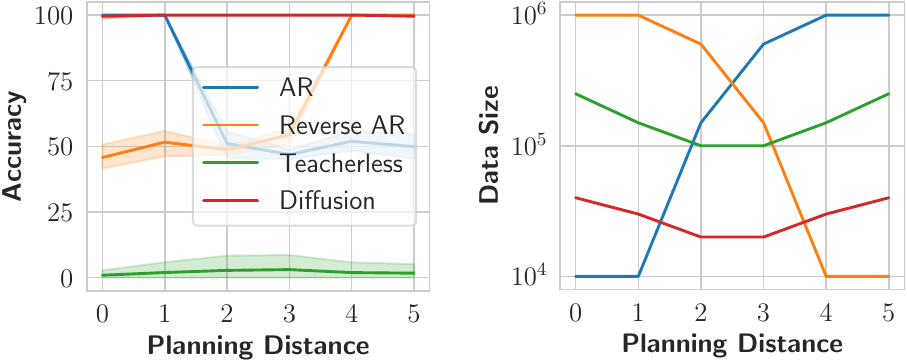}

\caption{\textbf{(Left)} Accuracy of different method given 50k training data. \textbf{(Right)} Minimum data size required to solve (i.e., accuracy above 90\%) subgoal at each planning distance.}
\label{fig:findpath-PD}
\end{minipage}

\vspace{-5pt}
\end{figure} 
We designed a simple planning task to serve as our running example. Consider the example in Figure~\ref{fig:findpath}, where the input for the task consists of a set of shuffled edges from the graph shown below. At the end of the input sequence, the start and goal nodes are specified to indicate the path the model needs to find. The objective of this task is to identify the correct path in the graph and output its constituent edges. The complexity of this problem arises from distracting factors (highlighted in orange) that potentially mislead the path selection. For instance, at node $7$, with the goal being node $9$, the model must plan over a distance of $3$ nodes to determine that the correct next choice should be node $5$ rather than $0$. We define this span as the Planning Distance (PD), a parameter adjustable in our synthetic task data. Intuitively, as the PD increases, the model faces greater difficulty in learning to determine the correct subsequent node. We formalize this intuition as \emph{subgoal imbalance}.

\begin{proposition} (\textbf{Subgoal imbalance due to the unknown data distribution $q(\rvx)$})
\reb{
Given the data $\rvx$ sampled from an unknown data distribution $q(\rvx)$, the difficulty of learning each subgoal $\vx_n$ can differ significantly based on how we parametrize the model distribution, and some subgoals may require substantially more data to learn or may even be infeasible to learn.}
\end{proposition}

\reb{\paragraph{Subgoal imbalance in autoregressive modeling.}
Given the data $\rvx$ sampled from an unknown data distribution $q(\rvx)$, the naive autoregressive modeling parametrizes the model distribution $p_\vtheta(\rvx)$ as $p_\vtheta(\vx_1)\prod_{n=2}^{N}p_\vtheta(\vx_n \mid \vx_{1:n-1})$, the difficulty of learning each subgoal $\vx_n$ can differ significantly as given only the left context, and some subgoals may require substantially more data to learn or may even be infeasible to learn.}
\paragraph{Setup.} We synthesize the data with only one distracting path. We randomize node numbers in $[0,10]$ and the intersection positions in $[0,5]$. We further designed this task to be symmetric, ensuring that simply training with reversed output, as suggested by~\citet{bachmann2024pitfalls}, cannot solve subgoals with all PDs. For comparison, we include Auto-regressive (AR), reverse AR~\citep{bachmann2024pitfalls}, and teacherless training~\citep{monea2023pass,bachmann2024pitfalls}, which can be seen as a lookahead method that produce all target tokens from the source input, and our proposed diffusion model (detailed in \S\ref{sec:how-diffusion}). For all the models, we keep the model architecture fixed as the same $3$-layer Transformer with approximately $6$M parameters. More details can be found in Appendix~\S\ref{app:exp}.

\paragraph{Discussion.} We examine the performance of all the models in two scenarios. In the first scenario, we generate a fixed number of 50k instances with mixed planning distance. We  plot the accuracy on the held-out evaluation set for each model in the left figure of Figure~\ref{fig:findpath-PD}.
Our findings indicate that autoregressive models (AR and Reverse AR) are only effective in solving cases where the PD equals 0 or 1 (or equivalently, 5 and 4 in the reverse setting). Due to the aforementioned subgoal imbalance phenomenon, when PD is less than 2, the task barely involves any planning, allowing models to simply copy from the input with ease. However, for larger PDs, AR models barely outperform random guessing. Teacherless training fails to adequately fit the training data, resulting in the production of illegal paths. In contrast, our diffusion model achieves perfect accuracy across all PD values.

In the second scenario, we investigate whether the challenging subgoals can be naturally resolved through data or model scaling, akin to the success observed in large language models~\citep{kaplan2020scaling,wei2022emergent}.
To investigate this question, we gradually increase the size of the dataset for each model with different PDs and plot the minimum data size required to solve the subgoal in the right figure of Figure~\ref{fig:findpath-PD}. We find that the autoregressive models (AR and Reverse AR) can learn the easy cases of PD equal to 0 and 1 (or equivalently, 5 and 4 in the reverse setting) with only 10k data points. However, exponentially larger amounts of data are required to address increasingly challenging subgoals. Both teacherless training and diffusion models exhibit a similar U-shaped curve in their performance. This similarity can be attributed to the fact that teacherless training can be conceptualized as a special case of diffusion without an iterative noising and denoising process. In these models, solving edge PDs necessitates slightly more data. We hypothesize that this phenomenon occurs because the distance to other positions is shorter from the middle position (i.e., higher closeness centrality), thus providing the middle position with more nearby tokens to aid in prediction. Overall, autoregressive models require significantly more data to address all PDs compared to diffusion models, highlighting their relative data inefficiency.

In addition to our previous experiments, we conducted a series of tests to examine the effect of increasing the parameter count in autoregressive models while maintaining a fixed dataset size of 50,000 instances. Our findings reveal that scaling the original 6 million parameter model to 85 million, 303 million, and 1.5 billion parameters fails to resolve all PDs. Only upon fine-tuning a substantially larger model, specifically the LLaMA 7B model~\citep{touvron2023llama}, did we observe successful resolution of all PD subgoals.

\subsection{Effective Hard Subgoal Learning in Diffusion Modeling}
\label{sec:how-diffusion}
These experiments collectively indicate that diffusion models are significantly more effective in learning challenging subgoals arising from subgoal imbalance. To elucidate why diffusion models exhibit this superior capability, we first establish a connection between autoregressive (AR) models and diffusion models by reformulating Equation (\ref{eq:dm_original}).
Instead of evaluating the KL divergence between two complicated categoricals~\citep{hoogeboom2021argmax}, we consider discrete diffusion with absorbing state and simplify it as the weighted cross-entropy losses~\citep{austin2021structured,Zheng2023ARD,shi2024simplified,sahoo2024simple}:
\begin{equation}
    D_{\mathrm{KL}}[q(\rvx_{t-1} | \rvx_t, \rvx_0) \vert\vert 
        p_{\vtheta}(\rvx_{t-1}|\rvx_t)=-w(t)\sum_{n=1}^N \1_\mathrm{\rvx_{t,n}\neq\rvx_{0,n}} \rvx_{0,n}^\top \log f(\rvx_{t};\vtheta)_n, 
\end{equation}
where $w(t)=\frac{\alpha_{t-1}-\alpha_{t}}{1-\alpha_t} \in (0,1]$ is a time-dependent reweighting term which places higher weight when $t$ approaching 0.
We then rewrite Equation (\ref{eq:dm_original}) as:
\begin{equation}
    L_{\mathrm{DM}} = \E_{q(\rvx_0)} \sum_{n=1}^N  \underbrace{\sum_{t=1}^{T}w(t)\mathbb E_{q(\rvx_t|\rvx_0)}u(\rvx_0,\rvx_t, n;\vtheta)}_{-\log p_{\text{DM}}(\vx_n \mid \vx_{\neq n})},
\label{eq:dm_rewrite}
\end{equation}
where $u(\rvx_0,\rvx_t, n;\vtheta) \coloneqq -\1_\mathrm{\rvx_{t,n}\neq\rvx_{0,n}} \rvx_{0,n}^\top \log f(\rvx_{t};\vtheta)_n$ is the cross entropy loss on token $n$. 
\begin{wrapfigure}{r}{0.35\textwidth}
    \vspace{-5pt}
    \begin{center}
    \includegraphics[width=0.99\linewidth]{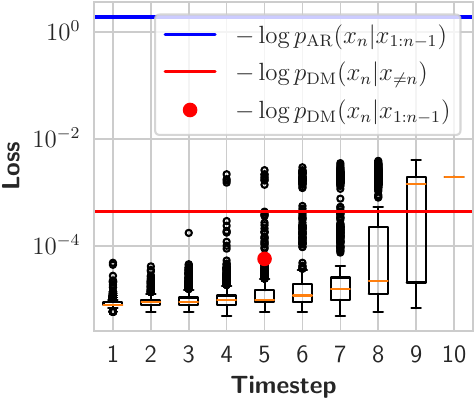}
    \end{center}
    \vspace{-10pt}
    \caption{Loss for a specific hard subgoal, i.e., PD$=$3, in Diffusion and AR modeling. We also show the unweighted loss $u(\rvx_0,\rvx_t, n;\vtheta)$ at different timestep $t$ and context $\rvx_t$ in diffusion modeling.}
    \label{fig:hard-token-loss}
    \vspace{-20pt}
\end{wrapfigure}

We can now systematically compare the losses of autoregressive (AR) and diffusion models (DM), specifically $-\log p_{\text{AR}}(\vx_n \mid \vx_{1:n-1})$ and $-\log p_{\text{DM}}(\vx_n \mid \vx_{\neq n})$, as expressed in Equations (\ref{eq:arloss}) and (\ref{eq:dm_rewrite}), respectively. In Figure~\ref{fig:hard-token-loss}, we examine a specific hard subgoal with Planning Distance (PD) equals 3 in both model types. The loss levels of AR and diffusion models are depicted using blue and red lines, respectively. The overall loss $-\log p_{\text{DM}}(\vx_n \mid \vx_{\neq n})$ in the diffusion model remains relatively low compared to its autoregressive counterpart $-\log p_{\text{AR}}(\vx_n \mid \vx_{1:n-1})$, corroborating the superior performance of the diffusion model on these challenging subgoals in our experiments.

Further analysis of the unweighted loss $u(\rvx_0,\rvx_t, n;\vtheta)$ in the diffusion model, based on 1,000 samples of $\rvx_t \sim q(\rvx_t|\rvx_0)$, reveals a clear trend: as the number of timesteps increases, resulting in more noise in $\rvx_t$, objectives in smaller timesteps (i.e., recovery from less noisy data) become significantly easier to learn. \reb{From a multi-view learning perspective~\citep{xu2013survey}, each $\rvx_t$ can be interpreted as a distinct view of $\rvx_0$, where each view provides different information about $\rvx_0$.} In the diffusion process, by exploring the consistency and complementary properties of different views offered by a diverse range of interrelated objectives $u(\rvx_0,\rvx_t, n;\vtheta)$, our findings suggest that objectives challenging to learn in AR models become more effective, promising, and exhibit better generalization in diffusion models.

This phenomenon is particularly evident when examining scenarios where mask noise is applied to positions after the hard token, i.e., $\rvx_t=\vx_{1:n-1}$, where the diffusion model learns the hard subgoal similarly to AR models. We plot this loss as $-\log p_{\text{DM}}(\vx_n \mid \vx_{1:n-1})$ in the figure. Unlike in the AR model, where this learning is consistently difficult, in diffusion models, this challenging subgoal is addressed at a much more manageable level during the learning process.

\subsection{Multi-granularity Diffusion Modeling}
\label{sec:mdm}
\reb{These observations provide valuable insights, i.e., diffusion modeling builds on a diverse range of interrelated views from the data $\rvx_0$ to handle a challenging subgoal. To handle multiple challenging subgoals in real data, we should prioritize different subgoals based on their difficulty during the learning process to achieve more effective learning outcomes and faster convergence, and this naturally translates to prioritizing difficult views as the learning of a subgoal depends on learning interrelated views related to it.} Building on this, we propose the multi-granularity diffusion model as a natural extension of the discrete diffusion model.

In practice, to optimize Equation (\ref{eq:dm_rewrite}), we typically employ Monte Carlo sampling, which results in:
\begin{equation}
L_{\mathrm{DM}} = \sum_{n=1}^N \sum_{t=1}^{T}w(t) u(\rvx_0,\rvx_t, n;\vtheta).
\label{eq:dm_mc}
\end{equation}
For a sequence of length $N$, the probability of sampling the same $\rvx_t$ in AR is 1. However, in diffusion, this probability reduces to $1/C_{N-1}^{t(N-1)/T}$ due to the randomness in sampling $\rvx_t$, potentially reducing the training efficiency of diffusion models. We note that Equation (\ref{eq:dm_mc}) employs a sequence-level reweighting term $w(t)$ to indicate the importance of $\rvx_t$. However, individual tokens within the sequence, given their imbalanced difficulties, are not properly reweighted. 
To address this, we propose multi-granularity diffusion modeling (\ourmodel), which introduces an additional token-level reweighting mechanism to enhance training efficiency:
\begin{equation}
\label{eq:mdm}
L_{\mathrm{\ourmodel}} = \sum_{n=1}^N \sum_{t=1}^{T}w(t) v(\rvx_{t,n})u(\rvx_0,\rvx_t, n;\vtheta),
\end{equation}

where $v(\rvx_{t,n}) = \alpha(1-\exp(-u(\cdot)))^\beta$ is the adaptive token-level reweighting term. Setting $\beta>0$ reduces the relative loss for easy tokens while emphasizing harder tokens, and $\alpha$ is used to control the relative reweighting magnitude. For inference, we employ an easy-first TopK decoding strategy, which has demonstrated superior performance compared to the random decoding method used by~\citet{austin2021structured}. This finding aligns with similar observations documented in prior studies~\citep{savinovstep2021,Zheng2023ARD}. We provide a detailed derivation and algorithm of the training and inference process in Appendix \S\ref{app-sec:discrete-details} and \S\ref{app:alg}, respectively.
\section{Experiments}
\label{sec:experiments}
In Section \S\ref{sec:planning_task} we show our model works well on a straightforward planning task with only one hard subgoal. However, it is important to note that real-world scenarios often involve instances with multiple challenging subgoals. In this section, we aim to assess the performance of our model in tackling three considerably more complex problem-solving tasks that necessitate deliberate planning. Detailed experimental setup can be found in Appendix~\S\ref{app:exp}.

\subsection{Countdown}
\label{sec:cd}
Countdown~\citep{countdown} is a mathematical reasoning challenge and is a generalized version of the game of 24, which even advanced models such as GPT-4 struggle with~\citep{yao2024tree}. The goal of Countdown is to use the given numbers and arithmetic operations ($+-*/$) to obtain a target number. For example, given 4 numbers ``97,38,3,17'' and a target number ``14'', a step-by-step solution is ``97-38=59,59-17=42,42/3=14''.

\paragraph{Setup.}
\begin{wraptable}{r}{0.5\textwidth}
\centering
\vspace{-20pt}
\caption{Results on the Countdown (CD) task with increasing complexity.}
\scalebox{0.86}{
\begin{tabular}{lcccc}
\toprule
& Params & CD 3 & CD 4 & CD 5 \\
\hline
\textit{Autoregressive}&&&&\\
\multirow{3}{*}{GPT-2 Scratch} & 6M & 94.1 & 31.9 & 4.3 \\
 & 85M & 95.9 & 45.8 & 5.1 \\
 & 303M & 96.4 & 41.3 & 4.5 \\
Stream-of-Search & 250M & - & 54.2 & - \\
\multirow{2}{*}{LLaMA} & 7B & 95.7 & 41.1 & 6.7 \\
 & 13B & 96.5 & 51.1 & 7.4 \\
\hline
\textit{Diffusion} &  &  & & \\
VDM & 85M & 99.1 & 73.4 & 16.3 \\
D3PM & 85M & 99.4 & 83.1 & 27.6 \\
RDM & 85M & 99.5 & 87.0 & 45.8 \\
\multirow{3}{*}{\textbf{\ourmodel(Ours)}} & 6M & 98.1 & 52.0 & 27.0 \\
 & 85M & 99.5 & \textbf{91.5} & \textbf{46.6} \\
 & 303M & \textbf{99.9} & 88.3 & 39.0 \\
 \bottomrule
\end{tabular}}
\label{tab:cd-main}
\vspace{-10pt}
\end{wraptable}
We follow ~\citet{gandhi2024stream} to generate 500k problems with target numbers ranging from 10 to 100 and randomly hold out 10\% of the targets for `out-of-distribution' evaluation. We consider three subtasks with increasing complexity by varying the number of input digits in \{3,4,5\}. Given that search-augmented prompting approaches~\citep{yao2024tree} have recently been employed to address the limitations of AR, we also compare with such approaches by training on Countdown 4 and evaluating on the same game of 24 test set as \citet{yao2024tree}.

\paragraph{Baselines.} Our primary comparison involves autoregressive models trained from scratch, employing the GPT-2 architecture~\citep{radford2019language} with parameter sizes ranging from 6M, 85M, and 303M (denoted as GPT-2 Scratch). We also include larger pre-trained AR models LLaMA~\citep{touvron2023llama} with sizes 7B and 13B. These models are fine-tuned using the same dataset. In addition, we compare with Stream-of-Search~\citep{gandhi2024stream}, which augments the dataset with search trajectory such that the AR model can be taught
to search. 
Furthermore, we compare with several existing diffusion models, both continuous models VDM~\citep{kingma2021variational} and discrete models D3PM~\cite{austin2021structured} and RDM~\citep{Zheng2023ARD}. By default, we use the absorbing noise for discrete diffusion as it significantly outperforms the multinomial one~\citep{austin2021structured,Zheng2023ARD}. 
Finally, we consider in-context learning~\citep{brown2020language,wu2023self,ye2023compositional} based on GPT-4, including vanilla input-output (IO), chain-of-thought (CoT;~\citealt{wei2022chain}), CoT with Self-consistency (CoT-SC;~\citealt{wangself}) and Tree-of-thought (ToT;~\citealt{yao2024tree}). We use 5 in-context examples following ~\citealt{yao2024tree}.

\paragraph{Results on Countdown.}
As shown in Table~\ref{tab:cd-main}, diffusion-based approaches demonstrate superior performance across all three Countdown tasks compared to autoregressive models, especially as the complexity of the tasks increases. 
We have several key findings based on the result. Firstly, the 6M diffusion model outperforms both the 303M GPT-2 model trained from scratch and the pretrained 13B LLaMA model, indicating that the modeling approach sometimes outweighs the sheer number of parameters. Secondly, while training with search trajectory supervision (Stream-of-search) does provide some benefits, its effectiveness is limited. Importantly, training the entire search trajectory as a sequence poses additional challenges due to its long length, such as in the case of Countdown 5 where the search trajectories can span 60,000 tokens. Lastly, our model surpasses all previous diffusion models, demonstrating the efficacy of the multigranularity loss.

\paragraph{Results on Game of 24.}
\begin{wraptable}{r}{0.33\textwidth}
\vspace{-5pt}
\captionof{table}{Accuracy and token cost on game of 24.}
\centering
\scalebox{0.88}{
\begin{tabular}{lcc}
\toprule
 & Acc. & \# Token \\
 \hline
 \multicolumn{3}{l}{\textit{Prompting}} \\
GPT-4 IO & 7.3 & x28 \\
GPT-4 CoT & 4.0 & x61 \\
GPT-4 CoT-SC & 9.0 & x241 \\
GPT-4 ToT & 74.0 & x186 \\
\multicolumn{3}{l}{\textit{Supervised training}} \\
GPT-2 Scratch & 18.8 & \textbf{x1} \\
\textbf{\ourmodel} & \textbf{76.0} & \textbf{x1} \\
\bottomrule
\end{tabular}}
\label{tab:tot24}
\end{wraptable}As shown in Table~\ref{tab:tot24}, the performance of the GPT-4 with IO, CoT, and CoT-SC prompting methods from ~\citet{yao2024tree} is unsatisfactory for the given task, with only accuracy below 10\%. The introduction of ToT, which incorporates a search algorithm designed by human experts into the decoding process, significantly enhances the performance of GPT-4. This integration allows the AR model to backtrack as needed, resulting in notable improvements. However, this paradigm requires the assessment of intermediate steps using LLM, resulting in considerable computational costs due to the need for multiple LLM calls. We list the token cost in Table~\ref{tab:tot24} with more details in Appendix \ref{app-sec:24game}. ToT consumes 186 times more tokens than \ourmodel, showcasing the `internal' search capability by promoting global consistency in diffusion modeling. 
In summary, our model, despite having a parameter size of only 85M, significantly outperforms both the AR task-specific model of the same size (GPT-2 Scratch) in terms of performance and the larger general pre-trained model (GPT-4) in computation cost, indicating it is challenging for model scaling and decoding strategies to substitute the advantages of modeling paradigm.

\subsection{Sudoku}
\begin{wrapfigure}{r}{0.65\textwidth}
    \vspace{-40pt}
    \begin{center}
    \includegraphics[width=1\linewidth]{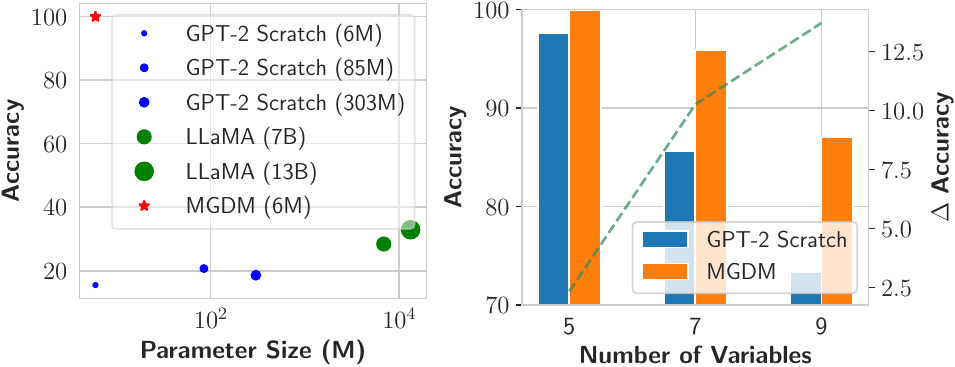}
    \end{center}
    \caption{\textbf{(Left)} Accuracy on Sudoku. \textbf{(Right)} Accuracy on boolean satisfiability problem with increasing difficulty.}
    \label{fig:sudoku-sat}
    \vspace{-20pt}
\end{wrapfigure}
\label{sec:sudoku}
Sudoku is a classical logic-based number placement puzzle that has gained popularity due to its rigorous intellectual demands.
The goal of Sudoku is to meticulously fill a 9 × 9 grid with numerical digits, ensuring that every column, row, and 3 × 3 subgrid contains all the numbers from 1 to 9.

\paragraph{Setup.}
We collect one million solved games from ~\citet{park2016sudoku} and use the first 100k as our training set and the subsequent 1k as the testing set. We employ the digit 0 to represent the vacant position that needs to be filled. We then transform the 9 × 9 grid into a sequence of 81 digits, which serves as the model input. To illustrate, an example input appears as ``080050060...603100007'' (omitted for brevity), while the corresponding output is represented as ``789251364...653184297''. During tokenization, we treat each digit as a separate token.
\paragraph{Result.}
We show the results in the left figure of Figure~\ref{fig:sudoku-sat}. As the size of the AR model increases, the performance remains unsatisfactory. For instance, the LLaMA model achieves a performance of only 32.9 with 13B parameters. In contrast, our model, which has only 6M parameters, is able to perfectly solve all the problems, demonstrating the significant advantage brought by the modeling architecture.

\subsection{Boolean Satisfiability Problem}
\label{sec:sat}
The Boolean satisfiability problem, commonly known as SAT, is a foundational problem in computer science that has been rigorously proven to be NP-complete~\citep{cook1971complexity}. This challenging combinatorial problem is attractive as a broad range of search problems from 
domains such as software verification, test pattern generation, planning, scheduling, and combinatorics can all routinely be solved by reducing to an appropriate SAT problem~\citep{gomes2008satisfiability}. 
The goal of SAT is to determine whether a given Boolean formula represented in conjunctive normal form (CNF) can be assigned a set of values (0 or 1) to its variables, such that the formula evaluates to true (1). An example formula with three variables can be $(x_1 \vee \neg x_2) \wedge (\neg x_1 \vee x_2 \vee x_3) \wedge \neg x_1$ and an corresponding assignment is $x_1=0, x_2=0, x_3=1$.
\paragraph{Setup.} Given the number of variables $n$ and clauses $m$, we iteratively sample $k=3$ variables (and their polarities) uniformly at random until $m$ clauses are obtained. To ensure that we get relatively hard instances of SAT, we take advantage of the well-studied family of random $k$-SAT problem~\citep{ding2015proof} and set the $m$ to be close to $m = 4.258n + 58.26n^{-2/3}$ given $n$, as it has been observed that SAT solvers are slow to determine the satisfiability of a formula when $m$ is near the threshold~\citep{crawford1996experimental}. We consider increasing numbers of variables from \{5,7,9\} and generate 50k training data for $n=5,7$ and 100k for $n=9$, as well as additional 1k testing data for each $n$. 
\paragraph{Result.}
As shown in the right figure of Figure~\ref{fig:sudoku-sat}, \ourmodel performs well in solving scenarios with five variables, while the AR model falls slightly short. As the number of variables increases, both our model and the AR model experience a certain degree of decrease in accuracy. However, the performance gap between the two models widens as the difficulty of the task increases. This indicates that our diffusion model exhibits a more pronounced advantage in handling more challenging tasks than the AR counterpart.

\subsection{Analysis}
\label{sec:analysis}
\begin{figure}[t]
  \centering
\begin{minipage}{.55\linewidth}
\scalebox{0.9}{
\begin{tabular}{lcc}
\toprule
 & Random & TopK \\
 \hline
No reweighting & 82.1 & 87.3 \\
\hline
Original sequence-reweighting & 83.1 & 88.5 \\
+ token-reweighting ($\alpha$=0.25, $\beta$=1) & 84.9 & \textbf{90.4} \\
+ token-reweighting ($\alpha$=1, $\beta$=1) & 82.4 & 89.3 \\
+ token-reweighting ($\alpha$=0.25, $\beta$=2) & 82.4 & 87.9 \\
\hline
Linear sequence-reweighting & 79.6 & 87.0 \\
+ token-reweighting ($\alpha$=0.25, $\beta$=1) & 83.2 & 88.0 \\
+ token-reweighting ($\alpha$=1, $\beta$=1) & 86.7 & 90.4 \\
+ token-reweighting ($\alpha$=0.25, $\beta$=2) & 85.6 & \textbf{91.5} \\
\bottomrule
\end{tabular}}
\vspace{5.5pt}
\captionof{table}[]{Ablation on training reweighting strategies and inference decoding methods.}
\label{tab:ablation}
\end{minipage}
\begin{minipage}{.33\linewidth}
\centering
\includegraphics[width=0.99 \textwidth]{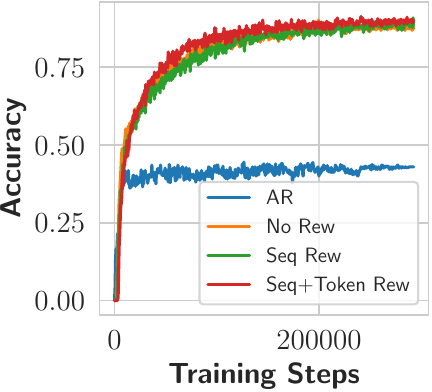}
\vspace{-10pt}
\caption{Evaluation accuracy throughout the training process for AR and \ourmodel with different reweighting strategies.}
\label{fig:convergence}
\end{minipage}
\vspace{-16pt}
\end{figure} 

\paragraph{On the Effect of Training and Decoding Strategies.}
As listed in Table~\ref{tab:ablation}, we find that changing the sequence-reweighting strategies has only led to a slight improvement in performance. However, when a suitable parameter is selected for token-reweighting, a more significant improvement can be observed. Additionally, the easy first decoding (TopK) outperforms the random one, which aligns with previous findings~\citep{ghazvininejad2019mask,Zheng2023ARD}. We compare the evaluation accuracy along the training process in Figure~\ref{fig:convergence}. By aligning the AR training steps with the diffusion process, we can see AR converges rapidly, with the performance tends to plateau afterward. The utilization of our multi-granularity loss, which incorporates sequence and token reweighting, demonstrates superior performance, particularly during the middle stages of training. This implies that the inclusion of such a loss function contributes to enhanced convergence during the training process.

\paragraph{On Decoding Speed.}
\begin{wrapfigure}{r}{0.6\textwidth}
    \vspace{-5pt}
    \begin{center}
    \includegraphics[width=0.99\linewidth]{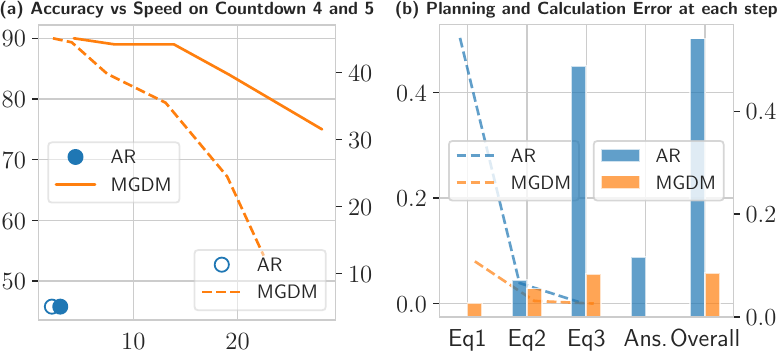}
    \end{center}
    \caption{\textbf{(a)} Accuracy and speed (samples per second) trade-off by varying the diffusion timesteps on Countdown 4 (left y-axis) and 5 (right y-axis).\textbf{(b)} Ratio of planning (left y-axis) and calculation (right y-axis) error at each reasoning step on Countdown 4.}
    \label{fig:analysis}
    \vspace{-5pt}
\end{wrapfigure}

We assess the trade-off between accuracy and decoding speed by comparing the performance of the AR model (GPT-2 Scratch 85M) and \ourmodel (85M). The speed metric is determined by the number of samples processed using a batch size of 1 on the NVIDIA GeForce V100 GPU. 
As shown in Figure~\ref{fig:analysis}(a), \ourmodel can flexibly control the trade-off between accuracy and decoding speed by varying the diffusion timesteps.
Notably, by employing just one diffusion step, \ourmodel demonstrates a remarkable 10x improvement in speed compared to AR, while maintaining superior accuracy with 75\% and 12.7\% compared to 45.8\% and 5.1\% of AR on Countdown 4 and 5, respectively.
We observed that the slope of countdown 4 is smaller compared to countdown 5 in the trade-off. This suggests that for tasks with lower complexity, diffusion demonstrates a more noticeable speed advantage by setting a smaller diffusion step. In addition, it also indicates that sacrificing some efficiency for performance improvements becomes particularly evident when dealing with more intricate tasks.

\paragraph{Error Analysis: The Regretful Compromise.}
To gain a deeper understanding of error patterns in AR and \ourmodel, we conducted an error analysis on Countdown 4. For instance, given the input ``7,38,3,1" and the target number 14, a correct solution would be ``97-38=59,59-17=42,42/3=14". We divide the solution into four parts: equation 1, equation 2, equation 3, and answer checking. 
First, from a calculation perspective, we assess the error ratio in each equation by comparing the left-hand side and the right-hand side, regardless of whether the correct number was chosen. 
As shown in Figure~\ref{fig:analysis}(b), the bar plot demonstrates that the majority of calculation errors for AR are concentrated in Equation 3. For example, given input ``16,4,40,51" and target 87, the prediction of AR is ``51-40=11,16*4=64,11+64=87" while the correct solution is ``16/4=4,40+51=91,91-4=87". 
\reb{The first two equations are calculated correctly, but the last one is incorrectly forced to equal 87. Moreover, in many cases, the error rate for the last equation in AR (48.9\%) is significantly higher than that of the previous equations (0.2\% for the first and 7.2\% for the second). This indicates that the model's lack of planning capability results in its realization of being unable to achieve the goal only at the end. By that point, it cannot correct the previous errors, leading it “regretfully” to rely on incorrect equations to reach the goal (e.g., the number 87 in the above case).}
This significantly increases the number of calculation errors in the third equation. We call this phenomenon `The Regretful Compromise'. 
The reason for this is that the AR model made incorrect choices of numbers or operations in previous equations. This is demonstrated by examining the step at which the models fail the task, as depicted in Figure~\ref{fig:analysis}(b). It is evident that there is a notable frequency of planning errors in the first equation for the AR model, where the number of errors is significantly higher compared to our model. This highlights the limitations of the left-to-right decoding approach in AR, which adversely affects its planning ability.

\section{Related Work}
\subsection{Autoregressive Modeling}
Starting from \citet{bengio2000neural} and later \citet{sutskever2011generating}, the autoregressive modeling paradigm, where the prediction of a token only depends on the preceding context, is widely adopted in modeling language, until recently~\citep[\emph{inter alia}]{chatgpt,achiam2023gpt,claude,team2023gemini,touvron2023llama,jiang2023mistral,bai2023qwen}. Theoretically, the autoregressive Transformers have limited expressive power, but their capabilities can be expanded given sufficient chain-of-thought intermediate steps~\citep{wei2022chain,merrill2023expresssive,malach2023auto}. However, ~\citet{lin2021limitations} demonstrates that the expressing of some next-tokens requires super-polynomial computational resources and is NP-hard to approximate.
Numerous advancements have been made upon the AR paradigm to compensate for modeling deficiencies, such as reverse training~\citep{lee2023teaching,golovneva2024reverse},fill-in-the-middle training~\citep{bavarian2022efficient}, future-token prediction~\citep{qi2020prophetnet,gloeckle2024better}, lookahead attention~\citep{du2023autoregressive} during the training stage, as well as search-augmented decoding~\citep[\emph{inter alia}]{lu2022neurologic,xie2023decomposition,yao2024tree} during inference. 
In practice, autoregressive next-token predictors are shown to be ill-suited for planning tasks~\citep{bubeck2023sparks, valmeekam2023planning,valmeekam2024planbench,dziri2024faith,kambhampati2024llms}.
Besides, ~\citet{bachmann2024pitfalls,lin2024rho} find not all tokens are equal and some tokens are hard to learn in the AR pretraining stage, implying the introduced subgoal imbalance phenomenon also exists in the general text corpus.

\subsection{Non-autoregressive Modeling}
The non-autoregressive (NAR) generation method, which produces all target tokens simultaneously given the source context, is first proposed by ~\citep{gu2017non} in the text field for machine translation, primarily due to the efficiency consideration. While a series of advancements have been made afterward~\citep[\emph{inter alia}]{lee2018deterministic,gu2019levenshtein,ghazvininejad2019mask,qian2021glancing,huang2022directed}, traditional NAR models still tend to underperform AR models in terms of generation quality~\citep{xiao2023survey}.
Diffusion models~\citep{sohl2015deep,ho2020denoising}, a powerful class of generative models known for their impressive image-generation capabilities~\citep{dhariwal2021diffusion}, have recently been applied to the field of text generation~\citep{hoogeboom2021argmax,austin2021structured,li2022diffusion,campbell2022continuous,dieleman2022continuous,chen2023analog,ye2023dinoiser,lovelace2024latent}, reinforcement learning~\citep{janner2022planning,chi2023diffusion} and protein design~\citep{xu2022geodiff,hoogeboom2022equivariant,corso2023diffdock}. In essence, diffusion models perform a multi-step denoising process to progressively convert a random noise into a data sample, and the denoising procedure can be seen as parameterizing the gradients of the data
distribution~\citep{song2019generative}, connecting them to score matching~\citep{hyvarinen2005estimation} and energy-based models~\citep{lecun2006tutorial}. 
For text, the diffusion model can be seen as an extension of the traditional iterative NAR models~\citep{gong2022diffuseq} and has been shown to approach or outperform AR models on text perplexity~\citep{han2023ssd,lou2023discrete,gulrajani2024likelihood}, diversity~\citep{gong2022diffuseq,diffuseqv2,zhang2023planner} as well as various seq-to-seq tasks~\citep{wu2023ar,Zheng2023ARD,ye2024diffusion}. 
In this work, we compare diffusion with AR from a perspective of subgoal imbalance and demonstrates the effectiveness of diffusion in tasks requiring complex reasoning and planning.

\section{Conclusion}
This paper presents an extensive analysis of the limitations of auto-regressive (AR) language models when applied to planning tasks that involve deliberate planning, both in controlled settings and real-world contexts.
Based on an advanced understanding, we propose an improved diffusion model, named \ourmodel, that performs significantly better than AR and previous diffusion models on various sophisticated planning tasks.
Our findings underscore the necessity to reevaluate the sequence modeling paradigm for modern large language models, especially in tackling challenging problem-solving tasks.

\subsubsection*{Acknowledgments}
This research was supported in part by the joint research scheme of the National Natural Science Foundation of China
(NSFC) and the Research Grants Council (RGC) under grant number N\_HKU714/21.

\bibliography{main}
\bibliographystyle{iclr2025_conference}
\newpage
\appendix
\section{More Background and Derivation of Discrete Diffusion Models}
\label{app-sec:discrete-details}
\subsection{Background}
Discrete diffusion probabilistic models are first introduced in \citet{sohl2015deep} for binary data, and later extended to categorical data in \citep{hoogeboom2021argmax}. \citet{austin2021structured} provides a general form of discrete diffusion and introduces multiple transition matrices, including an absorbing variant that draws close connections to masked language models \citep{devlin-etal-2019-bert}. 
Several subsequent works push this line of research further from various aspects, such as incorporating editing-based operations \citep{johnson2022beyond,reid2022diffuser}, casting permuted language models \citep{yang2019xlnet} as diffusion models \citep{hoogeboom2022autoregressive}, developing a continuous-time framework \citep{campbell2022continuous}, parameterizing the routing mechanism~\citep{Zheng2023ARD}, and investigating score functions for learning the reverse process \citet{sun2023score,lou2023discrete}.

\subsection{Derivation Setup}
We now provide a detailed derivation of the loss in Equation (\ref{eq:dm_rewrite}).
For a clear illustration, we initiate derivation with a single random variable $\vx_0$ and ultimately link it with the multi-variable sequence $\rvx_0$.
Suppose $\vx_0 \sim q(\vx_0)$ is a discrete random variable with $K$ possible categories and represented as a one-hot vector. The forward process $q(\vx_{1:T}|\vx_0) = \prod_{t=1}^T q(\vx_t|\vx_{t-1})$ corrupts the original data $\vx_0$ into a sequence of increasingly noisy latent variables $\vx_{1:T}\coloneqq \vx_1, \dots, \vx_T$. The learned backward process $p_{\vtheta}(\vx_{0:T})=p(\vx_T)\prod_{t=1}^Tp_{\vtheta}(\vx_{t-1}|\vx_t)$ gradually denoises the latent variables
to the data distribution. 
In discrete diffusion, both the forward and backward distribution are defined as categorical distribution, e.g., $q(\vx_t|\vx_{t-1})= \mathrm{Cat}(\vx_t;\vp = \mQ_t^\top\vx_{t-1})$ and $p_{\vtheta}(\vx_{t-1}|\vx_t)=q(\vx_{t-1} | \vx_t, f(\vx_t;\vtheta))$, where $\mQ_t$ is a pre-defined transition matrix of size $K \times K$~\citep{hoogeboom2021argmax,austin2021structured}.

\subsection{The marginal and posterior}
Starting from $\vx_0$, we obtain the following $t$-step marginal and posterior at time $t-1$:
\begin{align}
q(\vx_t | \vx_0) = \mathrm{Cat}\left(\vx_{t}; \vp = \overline{\mQ}_{t}^\top\vx_{0}  \right), 
    \quad \text{with}\quad 
  \overline{\mQ}_{t} = \mQ_1  \mQ_2 \hdots \mQ_t \nonumber \\
 q(\vx_{t-1}|\vx_{t}, \vx_0) = \frac{q(\vx_{t}|\vx_{t-1}, \vx_0)q(\vx_{t-1}|\vx_0)}{q(\vx_{t}| \vx_0)}
    =\mathrm{Cat}\left(\vx_{t-1};\vp = \frac{\mQ_t\vx_t \odot \overline{\mQ}_{t-1}^\top \vx_0   }{\vx_t^\top\overline{\mQ}_{t}^\top \vx_0}\right),
    \label{eq:posterior-original}
\end{align}
where $q(\vx_t|\vx_{t-1}, \vx_0)= q(\vx_{t}|\vx_{t-1})$ due to the Markov property of the forward process. The KL divergence between $q$ and $p_\vtheta$ can be computed by simply summing over all possible values of each random variable.
The cumulative products $\overline{\mQ}_t$, which can be computed in closed form or precomputed for all $t$ depending on the choice $\mQ_t$, may be prohibitive for large $T$ and number of categories. Therefore, two commonly used forms of $\mQ$ are introduced by ~\citet{hoogeboom2021argmax} and ~\citet{austin2021structured}, which ensures $\overline{\mQ}_t$ can still be computed efficiently, allowing the framework to scale to a larger number of categories.

\subsection{Transition Matrix}
\citet{austin2021structured} introduced multiple types of the transition matrix $\mQ_t$, such as uniform~\citep{hoogeboom2021argmax}, absorbing, discretized Gaussian and token embedding distance.
The absorbing noise for discrete diffusion has been demonstrated to outperform the others~\citep{austin2021structured}, where the transition matrix is given by :
\begin{align*}
    \left[\mQ_t\right]_{ij} = \begin{cases}
    1  \quad&\text{if}  \quad i = j = m \\
    1 - \beta_t \quad &\text{if} \quad i = j \ne m \\
    \beta_t \quad &\text{if} \quad j = m, i \ne m
\end{cases}.
\end{align*}
The transition matrix can also be written as $(1 - \beta_t) I + \beta_t \1 e^\top_m$, where $e_m$ is a vector with a one on the absorbing state $m$ and zeros elsewhere.
Since $m$ is an absorbing state, the corruption process converges not to a uniform distribution but to the point-mass distribution on $m$.
The transition matrices $\overline{\mQ}=\mQ_1  \mQ_2 \hdots \mQ_t$ can be computed in closed form. 
Specifically, we transition to another token with probability $\beta_t$ and stay the same with probability $1 - \beta_t$ in each step. After $t$ steps, the only operative quantity is the probability of not yet having transitioned to another token, given by ${\alpha_t} = \prod_{i=0}^t (1 - \beta_i)$. Therefore, we have $\overline{\mQ}_t={\alpha_t} I + (1 - {\alpha_t}) \1 e_m^\top$.

\subsection{Derivation of ELBO}
In order to optimize the generative model $p_{\vtheta}(\vx_{0})$ to fit the data distribution $q(\vx_{0})$, we typically
minimize a variational upper bound on the negative log-likelihood, defined below:
\begin{align*}
&-\log p_{\vtheta}(\vx_0) \\
&= -\log \int p_{\vtheta}(\vx_0, \vx_1, \dots, \vx_T) d\vx_1\cdots d\vx_T \\
&= -\log \int \frac{p_{\vtheta}(\vx_0, \vx_1, \dots, \vx_T)}{q(\vx_1, \dots, \vx_T | \vx_0)}q(\vx_1, \dots, \vx_T | \vx_0) d\vx_1\cdots d\vx_T \\
&= -\log \E_{q(\vx_1, \dots, \vx_T | \vx_0)}\Big[{\frac{p_{\vtheta}(\vx_0, \vx_1, \dots, \vx_T)}{q(\vx_1, \dots, \vx_T | \vx_0)}}\Big]\\
&\leq -\E_{q(\vx_1, \dots, \vx_T | \vx_0)}\Big[{\log\frac{p_{\vtheta}(\vx_0, \vx_1, \dots, \vx_T)}{q(\vx_1, \dots, \vx_T | \vx_0)}}\Big]\\
&= -\E_{q(\vx_1, \dots, \vx_T | \vx_0)}\Big[{\log\frac{p_{\vtheta}(\vx_0|\vx_1)p_{\vtheta}(\vx_T)  \prod_{t = 2}^T p_{\vtheta}(\vx_{t-1} | \vx_{t})}{q(\vx_T | \vx_0) \prod_{t=2}^T q(\vx_{t-1} | \vx_{t} ,\vx_0)}}\Big]\\
&= -\E_{q(\vx_1, \dots, \vx_T | \vx_0)}\Big[{\log p_{\vtheta}(\vx_0|\vx_1) - \sum_{t=2}^T \log \frac{q(\vx_{t-1} | \vx_{t} ,\vx_0)}{p_{\vtheta}(\vx_{t-1} | \vx_{t})} - \log \frac{q(\vx_T | \vx_0)}{p_{\vtheta}(\vx_T)}}\Big]\\
&= -\E_q\Big[{\log p_{\vtheta}(\vx_0|\vx_1) - \sum_{t=2}^T \KL[{q(\vx_{t-1} | \vx_{t} ,\vx_0)}\vert\vert{p_{\vtheta}(\vx_{t-1} | \vx_{t})}] - \underbrace{D_{\mathrm{KL}}[q(\vx_T | \vx_0) \vert\vert p(\vx_T)]}_{L_T(\text{const})}} \Big]\\
&= \underbrace{-\E_{q(\vx_1|\vx_0)}{\log p_{\vtheta}(\vx_0|\vx_1)}}_{{L}_0} + \sum_{t=2}^T \underbrace{\E_{q(\vx_t|\vx_0)}\Big[{\KL[{q(\vx_{t-1} | \vx_{t} ,\vx_0)}\vert\vert{p_{\vtheta}(\vx_{t-1} | \vx_{t})}]}\Big]}_{{L}_{t-1}} + L_T(\text{const}).
\end{align*}

\subsection{Derivation for Equation (\ref{eq:dm_rewrite})}
\label{app-sec:simplified-loss}
The categorical distribution $q(\vx_{t-1} | \vx_t, \vx_0)$ based on Equation (\ref{eq:posterior-original}) is given as:
\begin{align*}
    &q(\vx_{t-1} | \vx_t, \vx_0) \\
    &= \frac{\mQ_t \vx_t \odot \overline{\mQ}_{t-1}^\top \vx_0}{\vx_t^\top\overline{\mQ}_{t}^\top \vx_0} \\
    &=\! \frac{\left[(1-\beta_t) \vx_t + \beta_t\sigma_{\vx_t}\1\right] \odot \left[\alpha_{t-1} \vx_0 + (1 - \alpha_{t-1}) e_m\right]}{\alpha_t \vx_t^\top \vx_0 + (1-\alpha_t)\vx_t^\top e_m} \\
    &=\! \frac{(1-\beta_t) \alpha_{t-1} {\vx_t \!\odot\! \vx_0} \!+\! (1-\beta_t) (1 \!-\! \alpha_{t-1}){\vx_t \!\odot\! e_m} \!+\! \beta_t \alpha_{t-1}\sigma_{\vx_t}\!{\1 \!\odot\! \vx_0} \!+\! \beta_t (1 \!-\! \alpha_{t-1})\sigma_{\vx_t}\!{\1 \!\odot\! e_m} }{\alpha_t \vx_t^\top \vx_0 + (1-\alpha_t){\vx_t^\top e_m}}\\
    &=\! \frac{(1-\beta_t) \alpha_{t-1} {\vx_t \!\odot\! \vx_0} + (1-\beta_t) (1 \!-\! \alpha_{t-1}){\sigma_{\vx_t}\vx_t} + \beta_t \alpha_{t-1}\sigma_{\vx_t}{\vx_0}  + \beta_t (1 \!-\! \alpha_{t-1})\sigma_{\vx_t}{e_m}}{\alpha_t \vx_t^\top \vx_0 + (1-\alpha_t){\sigma_{\vx_t}}},
\end{align*}
where $\sigma_{\vx_t} \coloneqq e_m(\vu=\vx_t)$ represents the probability of noise drawn from $e_m$ being equal to $\vx_t$. Note $\vx_t \odot \vx_0=0$ if $\vx_t \neq \vx_0$ otherwise 1. Thus the computation of $q(\vx_{t-1} | \vx_{t}, \vx_0)$ breaks down into two cases:
\begin{align*}
    q(\vx_{t-1} | \vx_t, \vx_0) = \begin{cases}
       \eta_t\vx_t + \left(1-\eta_t\right) e_m, &\text{ if } \vx_t = \vx_0 \\
       \lambda_t\vx_0 + \left(1-\lambda_t\right) e_m(\vx_t), &\text{ if } \vx_t \neq \vx_0,
    \end{cases}
\end{align*}
where $\eta_t \coloneqq 1 - \frac{\beta_t(1-\alpha_{t-1})e_m(\vu=\vx_t)}{\alpha_{t} + (1 - \alpha_{t})e_m(\vu=\vx_t)}$, $\lambda_t \coloneqq \frac{\alpha_{t-1} - \alpha_{t}}{1-\alpha_t}$, and $e_m(\vx_t) = (1-\beta_t) \vx_t + \beta_t e_m$ denotes a noise distribution that interpolates between $\vx_t$ and $e_m$.

Recall the distribution $p_{\vtheta}(\vx_{t-1}|\vx_t)$ is parameterized by $q(\vx_{t-1} | \vx_t, f(\vx_t;\vtheta))$, the KL divergence between $q(\vx_{t-1} | \vx_t, \vx_0)$ and $p_{\vtheta}(\vx_{t-1}|\vx_t)$ becomes 0 when $\vx_t =\vx_0$. 
In the case of absorbing diffusion, $\vx_t=e_m$ if $\vx_t \neq \vx_0$ and $e_m(\vx_t)=e_m$. $q(\vx_{t-1} | \vx_t, \vx_0)$ has probability $\lambda_t$ on index $x_0$ and $1-\lambda_t$ on the absorbing state. The model $f(\vx_{t};\vtheta)$ has zero-probability on the absorbing state as it never predicts the mask token. Therefore, $p_{\vtheta}(\vx_{t-1}|\vx_t)$ also has $1-\lambda_t$ probability on the absorbing state. Putting them together, we derive the KL divergence as:
\begin{align*}
    D_{\mathrm{KL}}[q(\vx_{t-1} | \vx_t, \vx_0) \vert\vert 
        p_{\vtheta}(\vx_{t-1}|\vx_t)]&=1_{x_{t}\neq x_{0}}[\lambda_t
         \log \frac{\lambda_t}{f(\vx_{t};\vtheta)_{x_0}}+(1-\lambda_t)\log\frac{1-\lambda_t}{1-\lambda_t}]\\
         &=-\lambda_t 1_{x_{t}\neq x_{0}}
         \vx_{0}^\top\log f(\vx_{t};\vtheta)+C,
\end{align*}
where $1_{x_{t}\neq x_{0}}$ is 1 if $x_{t}\neq x_{0}$ otherwise 0, and $C$ is a constant. Moreover, given $\alpha_0=1$ by definition and $\lambda_0=1$, we have: 
\begin{align*}
    L(\vx_0) = -\E_{q(\vx_0)}  \sum_{t=1}^{T} \lambda_t\mathbb E_{q(\vx_t|\vx_0)}1_{\vx_t\neq\vx_{0}} \vx_0^\top \log f(\vx_{t};\vtheta)
\end{align*}
for a single random variable, and  
\begin{align*}
    L(\rvx_0) = -\sum_{n=1}^N \E_{q(\rvx_{0,n})}  \sum_{t=1}^{T} \lambda_t\mathbb E_{q(\rvx_{t,n}|\rvx_{0,n})}1_{\rvx_{t,n}\neq\rvx_{0,n}} \rvx_{0,n}^\top \log f(\rvx_{t,n};\vtheta)
\end{align*}
for $\rvx_0$ that represents a sequence of random variables $\rvx_0=(\vx_{0,1}, \dots, \vx_{0,N})$, where the $\lambda_t$ also represents the reweighting term $w(t)$ in Equation (\ref{eq:dm_rewrite}).

\section{Algorithms for Training and Inference}
\label{app:alg}
The detailed algorithms for training and inference are illustrated in Algorithm 1 and 2, respectively. For conditional training and inference, we split $\rvx$ into $[\rvx^{\text{src}};\rvx^{\text{tgt}}]$ and freeze the condition part $\rvx^{\text{src}}$ during training and inference.


\begin{algorithm}[H] %
   \caption{Training \ourmodel}
   \label{alg:training}
    \begin{algorithmic}
    \State {\bfseries Input:} neural network $f\left(\cdot;\vtheta\right)$, data distribution $p_{\text{data}}(\rvx_{0,1:N})$, a custom sequence reweighting term $w(t)$, token reweighting parameters $\alpha$ and $\gamma$, timesteps $T$.
    \State {\bfseries Output:} model parameters $\vtheta$.
    \Repeat
        \State Draw $\rvx_{0,1:N} \sim p_{\text{data}}(\rvx_{0,1:N})$;
        \State Draw $t \in \operatorname{Uniform}(\{1,\dots,T\})$;
         \State Draw $\rvx_{t} \sim q(\rvx_{t} | \rvx_{0})$;
        \For{$n = 1,2,\dots,N$}
            \State Let $u(\rvx_0,\rvx_t, n;\vtheta) \coloneqq \1_\mathrm{\rvx_{t,n}\neq\rvx_{0,n}} \rvx_{0,n}^\top \log f(\rvx_{t};\vtheta)_n$;
            \State Let $v(\rvx_{t,n}) = \alpha(1-\exp u(\rvx_0,\rvx_t, n;\vtheta))^\gamma$;
        \EndFor
        \State $L_\vtheta = -w(t)\sum_{n=1}^N v(\rvx_{t,n})u(\rvx_0,\rvx_t, n;\vtheta)$;
        \State Minimize $L_\vtheta$ with respect to $\vtheta$;
    \Until{converged}
    \end{algorithmic}
\end{algorithm}

\begin{algorithm}[H] %
   \caption{Sampling from \ourmodel}
   \label{alg:inference}
    \begin{algorithmic}
    \State {\bfseries Input:} trained network $f\left(\cdot;\vtheta\right)$, mask token id $\vm$, timesteps $T$, temperature $\tau$.
    \State {\bfseries Output:} generated sample $\rvx_0$.
    \For{$n = 1,2,\dots,N$}
        \State Initialize $\rvx_{T,n} = \vm$;
    \EndFor
    \For{$t = T,\dots,1$}
    \State Define indicator $\rve_t=\operatorname{TopK}\left(f\!\left(\rvx_{t};\vtheta\right)\!\right)$ with indices in top-$t/T$ values set to 1 and others 0;
        \For{$n = 1,2,\dots,N$}
            \State Draw $\widetilde{\rvx}_{0,n} \sim \operatorname{Categorical}\left(f\!\left(\rvx_{t};\vtheta\right)\!/\tau\right)$;
            \State $\rvx_{t-1,n} = \rve_{t,n}\widetilde{\rvx}_{0,n} + (1-\rve_{t,n})\vm$;
        \EndFor
    \EndFor
    \State {\bfseries Return} $\rvx_{0,1:N}$.
    \end{algorithmic}
\end{algorithm}

\section{Additional Experimental Details}
\label{app:exp}
\subsection{Task Details}
\begin{table}[h]
\centering
\caption{Dataset statistics. Minimal and CD are short for the minimal planning task and Countdown, respectively.}
\scalebox{0.9}{
\begin{tabular}{lcccccccc}
\toprule
\textbf{} & \textbf{Minimal} & \textbf{CD3} & \textbf{CD4} & \textbf{CD5} & \textbf{Sudoku} & \textbf{3-SAT 5v} & \textbf{3-SAT 7v} & \textbf{3-SAT 9v} \\
\hline
Train Instance & 50k & 500k & 500k & 500k & 100k & 50k & 50k & 100k \\
Test Instance & 1k & 1k & 1k & 1k & 1k & 1k & 1k & 1k \\
Avg Input Token & 47 & 11 & 13 & 16 & 81 & 245 & 269 & 305 \\
Avg Output Token & 21 & 16 & 25 & 35 & 81 & 9 & 13 & 17 \\
Max Input Token & 49 & 12 & 15 & 18 & 81 & 245 & 269 & 305 \\
Max Output Token & 23 & 22 & 35 & 52 & 81 & 9 & 13 & 17 \\
\bottomrule
\end{tabular}}
\label{tab:task-statistics}
\end{table}

We show the statistics and input-output examples on each dataset in Table~\ref{tab:task-statistics} and Table~\ref{tab:tasks}, respectively.

\begin{table}[ht]
\centering
\caption{Model parameters with varying model size.}
\begin{tabular}{lccc}
\toprule
\textbf{} & \textbf{Tiny} & \textbf{Base} & \textbf{Medium} \\
\hline
Parameters & 6M & 85M & 303M \\
Num of Layer & 3 & 12 & 24 \\
Num of Head & 12 & 12 & 16 \\
Hidden Dim & 384 & 768 & 1024 \\
\bottomrule
\end{tabular}
\label{tab:model-args}
\end{table}
\subsection{\ourmodel Implementation Details}
We conduct all the experiments on NVIDIA V100-32G GPUs, and we use 8 GPUs for training and sampling. 
We mainly consider comparing diffusion and AR models trained from scratch with different model sizes, with arguments for each size listed in Table~\ref{tab:model-args}. We use the GPT-2 architecture for both \ourmodel and AR.
We set the learning rate to 1e-3 for the tiny model and 3e-4 for others, and we set the batch size to 1024 across all the models and tasks.
We train \ourmodel for 1200 epochs on the minimal planning task, 300 epochs on Sudoku, and 600 epochs on other datasets.
By default, we set the diffusion sampling steps to $T=20$ for tasks with average output tokens larger than 20, otherwise $T=10$. We use a decoding temperature $\tau=0.5$ for all tasks.
For all the experiments, we have verified the statistical significance by running them multiple times.

\subsection{Baseline Implementation Details}
We train the AR model until convergence, and the number of training epochs is set to 200 for the minimal planning task, 300 for SAT, and 40 for others. We keep other parameters, e.g., batch size and learning rate, the same as training \ourmodel.

For LLaMA~\citep{touvron2023llama}, we use LoRA fine-tuning~\citep{hulora2021} with lora rank setting to 16. We use a learning rate of 1e-4, a batch size of 256, and train for a maximum of 20 epochs to ensure the model has converged.
For GPT-4, we borrow the numbers from~\citet{yao2024tree}.

For all the diffusion baselines, we use the same transformer architecture as GPT-2 to control the variables. We set the training parameters the same as \ourmodel, e.g., number of training epochs to 600, learning rate to 3e-4, and batch size to 1024. During inference, we set decoding timesteps to 20 for all diffusion models as we didn't observe a clear performance improvement when scaling to 1024. 

\section{Additional Experiments}
\subsection{Token Consumption on Game of 24}
\label{app-sec:24game}
We show the detailed accuracy and token consumption on the game of 24 in Table~\ref{tab:24game-full}.

\begin{table}[ht]
\caption{Detailed accuracy and token consumption on game of 24.}
\centering
\begin{tabular}{lccc}
\toprule
\textbf{} & Accuracy & Prompt Tokens & {Generate Tokens} \\
\hline
GPT-4 IO & 7.3 & 1k & 18 \\
GPT-4 CoT & 4.0 & 2.2k & 67 \\
GPT-4 CoT-SC & 9.0 & 2.2k & {6.7k} \\
GPT-4 ToT & 74.0 & 1.4k & {2.5k} \\
GPT2-Scratch & 18.8 & 11 & 26 \\
\ourmodel & 76.0 & 11 & 26 \\
\bottomrule
\end{tabular}
\label{tab:24game-full}
\end{table}

\reb{\subsection{AR with token reweighting}
\label{app-sec:ar-token-reweighting}
We show the accuracy of AR with the same token reweighting mechanism in Equation~\ref{eq:mdm} on the minimal planning task in Table~\ref{tab:ar-reweighting}. We find that applying token reweighting to AR models still cannot solve subgoals with PD larger than 1 (i.e., with accuracy around 50\%), similar to the original AR. 

\begin{table}[h]
\centering
\caption{Results of AR with token reweighting.}
\begin{tabular}{lcc}
\toprule
{Planning Distance} & {AR} & {AR with token reweighting} \\ \hline
0 & 100 & 100 \\
1 & 100 & 100 \\
2 & 51.1 & 52.1 \\
3 & 46.9 & 51.5 \\
4 & 52.0 & 50.3 \\
5 & 49.9 & 51.9 \\
\bottomrule
\end{tabular}
\label{tab:ar-reweighting}
\end{table}}

\newpage
\reb{\subsection{Scaling both Data and Model Size}
\label{app-sec:scaling}
As an extension of Table~\ref{tab:cd-main}, we show the accuracy of AR and \ourmodel when both data and model size are increased in Table~\ref{tab:scaling}. We find scaling both data and model size is effective for both AR and \ourmodel.

\begin{table}[h]
\centering
\caption{Results of scaling both model and data size.}
\begin{tabular}{lll}
\toprule
 & AR & \ourmodel \\
 \hline
85M model, 500k data & 45.8 & 91.5 \\
303M model, 500k data & 41.3 & 88.3 \\
303M model, 1M data & 53.3 & 95.6 \\
\bottomrule
\end{tabular}
\label{tab:scaling}
\end{table}}

\subsection{Case Study}
\begin{table}[ht]
\centering
\caption{Example predictions on Countdown 4. For each sub-equation, we mark the planning error in \markred{red} and the calculation error in \textbf{bold}. AR exhibits more calculation errors in the last equation due to incorrect planning in the previous steps.}
\scalebox{0.8}{
\begin{tabular}{lrlll}
\toprule
\textbf{Numbers} & \multicolumn{1}{l}{\textbf{Goal}} & \textbf{Groundtruth} & \textbf{AR Prediction} & \textbf{\ourmodel Prediction} \\
\hline
64,36,52,42 & 14 & 64-52=12,36/12=3,42/3=14 & \markred{\textbf{64/36=2},52/2=26,\textbf{42-26=14}} & 64-52=12,36/12=3,42/3=14 \\
9,73,99,75 & 81 & 75-73=2,9*2=18,99-18=81 & \markred{99+75=174,\textbf{174/9=16,73+16=81}} & 75-73=2,9*2=18,99-18=81 \\
2,52,20,73 & 57 & 52-20=32,32/2=16,73-16=57 & \markred{2*20=40,73-52=21,\textbf{40+21=57}} & 52-20=32,32/2=16,73-16=57 \\
9,80,4,5 & 89 & 9+80=89,5-4=1,89*1=89 & \markred{9-5=4,4/4=1,80+1=81} & 9+80=89,5-4=1,89*1=89 \\
65,2,61,22 & 96 & 65-61=4,2+22=24,4*24=96 & 65-61=4,\markred{22*4=88,2+88=90} & 65-61=4,2+22=24,4*24=96 \\
42,47,9,14 & 81 & 47-42=5,14-5=9,9*9=81 & 47-42=5,\markred{14*5=70,\textbf{9+70=89}} & 47-42=5,\markred{14*5=70,\textbf{9+70=81}} \\
41,4,48,20 & 96 & 41*4=164,48+20=68,164-68=96 & \markred{48-41=7,20-7=13,\textbf{4*13=92}} & \markred{4*20=80,\textbf{41-40=2},48*2=96} \\
21,36,3,42 & 39 & 42-36=6,3*6=18,21+18=39 & \markred{36-21=15,15/3=5,42-5=37} & \markred{42-21=21,36/3=12,\textbf{21-12=39}}\\
\bottomrule
\end{tabular}}
\label{tab:morecases}
\end{table}
We show more prediction cases of the autoregressive model and our model in Table~\ref{tab:morecases}.

\begin{table}[t]
\caption{Task details by showing example input and output for each dataset.}
\centering
\begin{tabular}{l|p{6cm}|p{4cm}}
\toprule
\textbf{Task} & \textbf{Input Example} & \textbf{Output Example} \\
\hline
Minimal Planning & 2,10/10,4/11,5/2,0/8,2/0,11/6,2/1,9/5,3/4,1-8,3 & 8,2/2,0/0,11/11,5/5,3 \\
\hline
Countdown 3 & 15,44,79,50 & 44-15=29,79-29=50 \\
\hline
Countdown 4 & 86,28,13,31,96 & 86+28=114,31-13=18,114-18=96 \\
\hline
Countdown 5 & 50,36,82,44,31,51 & 44-36=8,82*31=2542,
8+2542=2550,2550/50=51 \\
\hline
Sudoku & 080050060\newline
460907108\newline
005000029\newline
970006500\newline
000872031\newline
300049000\newline
004025003\newline
010000480\newline
603100007 & 
789251364\newline
462937158\newline
135468729\newline
978316542\newline
546872931\newline
321549876\newline
894725613\newline
217693485\newline
653184297 \\
\hline
3-SAT 5v & 1,4,5/1,-4,-5/2,-4,5/-1,-2,5/3,4,5/-2,-4,-5/2,3,-4/-2,-3,5/1,2,4/1,-2,3/-1,3,5/1,-2,-4/1,4,-5/1,-2,-5/1,2,-5/-1,-3,-4/-1,3,-5/-1,3,4/2,-4,-5/-1,-4,5/1,-3,-5/1,3,-5/1,-3,-4/-2,3,5/1,2,5/-1,2,-4/1,-2,4/1,-4,5/3,4,-5/-1,2,-3/1,-3,5/-2,4,5/1,-2,5/-1,2,5/1,3,-4/-1,-4,-5/-2,-3,-4/2,4,5/-2,3,-4/-3,4,5/2,-3,5 & 1,2,3,-4,5 \\
\hline
3-SAT 7v & -2,-3,-7/2,-4,-7/-3,4,-5/1,2,-3/1,5,-7/-5,-6,-7/2,-5,6/2,-5,-6/-3,-4,6/-1,2,-4/-3,6,7/-2,-5,6/2,3,-7/-1,2,3/-2,3,-4/-1,3,7/1,-2,-7/2,4,6/1,2,-7/2,-3,-6/1,-2,6/-1,5,7/3,-6,-7/2,6,7/-2,-6,-7/-2,3,-5/3,5,-6/-2,6,-7/-1,-2,-7/5,-6,-7/2,-6,-7/-2,5,7/-3,-4,5/2,3,-4/-3,5,-7/3,-4,5/-2,3,-6/1,2,-6/1,4,-7/1,4,7/2,4,5/1,5,-6/1,3,4/2,3,7/1,-2,4 & 1,2,3,4,5,6,-7 \\
\hline
3-SAT 9v & 3,-4,-6/1,3,5/2,-7,8/1,-3,6/2,-3,-8/-4,-5,-7/1,-6,-9/1,8,-9/2,3,-9/3,-5,9/-3,7,9/-2,-3,9/-1,-5,-9/-2,-7,-9/-1,3,5/2,-5,-9/4,-7,-9/-2,3,-8/2,3,7/2,-4,6/-2,3,5/-2,-6,-8/-3,-4,-8/-2,6,7/-3,4,6/-3,-6,9/2,7,-9/2,4,-5/-3,-5,8/-4,5,-7/-4,-6,-8/2,-6,9/2,-5,9/1,4,-9/5,8,9/1,-6,7/-3,6,-9/1,4,-5/4,-6,9/-1,2,6/1,-2,-5/1,-2,-9/-4,7,9/-1,-4,-7/-3,5,-8/-1,-3,6/-2,-3,6/-3,6,9/-1,-5,8/1,-5,-9/1,4,8 & 1,2,3,4,-5,6,-7,-8,9 \\
\bottomrule
\end{tabular}\
\label{tab:tasks}
\end{table}

\end{document}